# A New Local Adaptive Thresholding Technique in Binarization

T.Romen Singh[1] , Sudipta Roy[2], O.Imocha Singh[3], Tejmani Sinam[4], Kh.Manglem Singh[5] .

[1]**Research Scholar**
**Department of Information Technology, School of Technology, Assam University,**
**Silchar – 788011, Assam, India.**

[2]**Associate Professor & Head,**
**Department of Information Technology, School of Technology, Assam University,**
**Silchar – 788011, Assam, India.**

[3]**Associate Professor & Head,**
**Department of Computer Science, Manipur University , Canchipur**
**Imphal – 795003, Manipur, India.**

[4]**Associate Professor ,**
**Department of Computer Science, Manipur University , Canchipur**
**Imphal – 795003, Manipur, India.**

[5]**Associate Professor**
**Department of Computer Science and Engineering, NIT, Manipur**
**Imphal -795001, Manipur, India.**

## Abstract

Image binarization is the process of separation of pixel values into two groups, white as background and black as foreground. Thresholding plays a major in binarization of images. Thresholding can be categorized into global thresholding and local thresholding. In images with uniform contrast distribution of background and foreground like document images, global thresholding is more appropriate. In degraded document images, where considerable background noise or variation in contrast and illumination exists, there exists many pixels that cannot be easily classified as foreground or background. In such cases, binarization with local thresholding is more appropriate. This paper describes a locally adaptive thresholding technique that removes background by using local mean and mean deviation. Normally the local mean computational time depends on the window size. Our technique uses integral sum image as a prior processing to calculate local mean. It does not involve calculations of standard deviations as in other local adaptive techniques. This along with the fact that calculations of mean is independent of window size speed up the process as compared to other local thresholding techniques.

## 1. Introduction

In general, scanned documents including text, line-drawings and graphics regions can be considered as mixed type documents. In many practical applications, we need to recognize or improve mainly the text content of the documents. In such cases, it is preferable to convert the documents into a binary form. Thus document binarization is the first step in most document analysis systems. The goal of document binarization is to convert the given input grayscale or color document into a bi-level representation. This representation is particularly convenient because most of the documents that occur in practice have one colour for text (e.g. black), and a different colour (e.g. white) for background.

Thresholding becomes a simple but effective tool to separate objects from the background. The output of the thresholding operation is a binary image whose one state will indicate the foreground objects, that is, printed text, a legend, a target, defective part of a material, etc., while the complementary state will correspond to the background. Depending on the application, the foreground can be represented by gray-level 0, that is, black for text, and the background by the highest luminance for document paper, that is, 1 in 8-bit images as white, or conversely the foreground by white and the background by black.

The binarization techniques for grayscale documents can be grouped into two broad categories: global binarization and local binarization. Global binarization methods such as one proposed by Otsu [8] try to find a single threshold value for the whole document. Then each pixel is assigned to page foreground or background based on its gray value. Global binarization methods are very fast and they give good results for typical scanned documents. For many years, the binarization of grayscale documents was based on the global thresholding algorithms [8]–[13]. These statistical methods, which can be considered as clustering approaches, are suitable for converting any grayscale image into a binary form but are inappropriate for complex documents, and even more, for degraded documents. If the illumination over the document is not uniform, for instance in the case of scanned book pages or camera-captured documents, global binarization methods tend to produce marginal noise along the page borders. To overcome these complexities, local thresholding techniques have been proposed for document binarization. These techniques estimate a different threshold for each pixel according to the grayscale information of the neighbouring pixels. The techniques of Bernsen [14], Chow and Kaneko [15], Eikvil [16], Mardia and Hainsworth [17], Niblack [18], Taxt [19], Yanowitz and Bruckstein [20] and Sauvola and Pietikainen [21] belong to this category. The hybrid techniques: O'Gorman [22] and Liu and Li [23], which combine information of global and local thresholds belong to another category.

In this paper, we focus on the binarization of grayscale documents using local adaptive thresholding technique, because in most cases colour documents can be converted to grayscale without losing much information as far as distinction between page foreground (text) and background is concerned. Some exceptions to this case are advertisements and some magazine styles. Local binarization methods [14]–[22] try to overcome these problems by computing thresholds individually for each pixel using information from the local neighborhood of the pixel. These methods are usually able to achieve good results even on severely degraded documents, but they are often slow since the computation of image features from the local neighborhood is to be done for each image pixel. This paper presents a fast approach to compute local thresholds without compromising the performance of local thresholding techniques using the technique of integral sum image as prior process for finding local mean of the neighbouring pixels in a window irrespective of window size. Using this approach we are able to achieve binarization speed close to the global binarization methods with the performance as good as that of local binarization schemes of Sauvola and Pietikainen [7],[21].

## 2. Integral Sum Image

The concept of integral sum images was popularized in computer vision by Viola and Jones [9] based on prior work in graphics [27]. An integral sum image *g* of an input image *I* is defined as the image in which the intensity at a pixel position is equal to the sum of the intensities of all the pixels above and to the left of that position in the original image.

Integral image intensity at (*x,y*) can be calculated as:

$$g(x, y) = \sum_{i=1}^{x} \sum_{j=1}^{y} I(i, j) \qquad (1)$$

The integral sum image of any grayscale image can be efficiently computed in a single pass as:

$$g(1, y) = I(1, y) + g(1, y-1), \quad y=2..n \qquad (2)$$

[integral sum of 1st row at (1,*y*)]

$$g(x, 1) = I(x, 1) + g(x-1, 1), \quad x=2..m \qquad (3)$$

[integral sum of 1st column at (*x*,1)]

$$g(x, y) = I(x, y) + g(x, y-1) + g(x-1, y) - g(x-1, y-1), \quad x=2..m, \quad y=2..n \qquad (4)$$

[ integral sum of entire pixels at (*x,y*)]

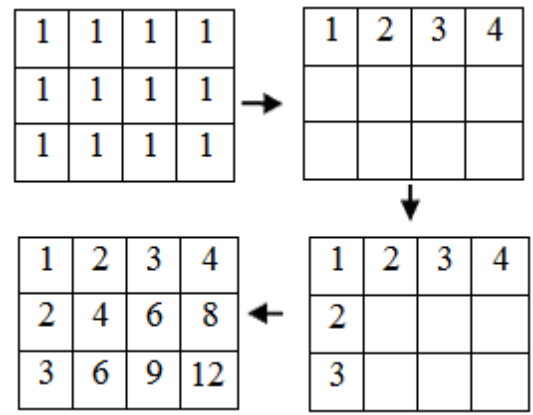

Fig. 1 Steps of integral sum calculation.

Thus the integral sum image *g* can be calculated using Eq.(2-4) in a single pass. Its diagrammatic sequence is shown in figure 1.

The local sum *s(x,y)* at (*x,y*) which is the centre of the local window of size *w*×*w* of an image *I* is the sum of all the pixel intensities within the local window. The sum $s(x,y)$ can be calculated in two passes as:

$$s(x,y) = \sum_{i=x-c}^{x+c} \sum_{j=y-c}^{y+c} I(i,j) \qquad (5)$$

where $c = \dfrac{w-1}{2}$, since $w$ is an odd number.

Once we have the integral sum image $g$, the local sum $s(x,y)$ of any window size can be computed simply by using two addition and one subtraction operations as in Eq.(6) unlike in Eq.(5) in one pass without depending on window size as:

$$s(x,y) = [g(x+d-1, y+d-1) + g(x-d, y-d)] \\ - [g(x-d, y+d-1) + g(x+d-1, y-d)] \qquad (6)$$

where $d = round(\dfrac{w}{2})$.

The local arithmetic mean $m(x,y)$ at $(x,y)$ is the average of the pixels within the window of size $w \times w$ of the image $I$. It can be calculated using Eq.(6) as:

$$m(x,y) = \dfrac{s(x,y)}{w^2} \qquad (7)$$

In this way the local mean can be calculated efficiently in a single pass without depending on local window size using integral sum image.

## 3. Locally Adaptive Thresholding

A threshold $T(x,y)$ is a value such that

$$b(x,y) = \begin{cases} 0 & \text{if } I(x,y) \leq T(x,y) \\ 1 & \text{otherwise} \end{cases} \qquad (8)$$

where $b(x,y)$ is the binarized image and $I(x,y) \in [0,1]$ be the intensity of a pixel at location $(x,y)$ of the image I. In local adaptive technique, a threshold is calculated for each pixel, based on some local statistics such as range, variance, or surface-fitting parameters of the neighborhood pixels. It can be approached in different ways such as background subtraction [29], water flow model [30], mean and standard derivation of pixel values [7], and local image contrast [31]. Some drawbacks of the local thresholding techniques are region size dependant, individual image characteristics, and time consuming. Therefore, some researchers use a hybrid approach that applies both global and local thresholding methods [32] and some use morphological operators[33]. Niblack [18], and Sauvola and Pietaksinen [7] use the local variance technique while Bernsen[14] uses midrange value within the local block.

### 3.1 Local variance methods

Niblack [18], and Sauvola and Pietaksinen [7] use the local variance technique. In these methods, the threshold is calculated based on the local mean $m(x,y)$ and standard deviation $\delta(x,y)$ within a window of size $w \times w$. Sauvola and Pietaksinen's method is an improvement on the Niblack's method, especially for stained and badly illuminated documents.

#### 3.1.1 Niblack's Technique

In this method the local threshold value $T(x,y)$ at $(x,y)$ is calculated within a window of size $w \times w$ as:

$$T(x,y) = m(x,y) + k\delta(x,y) \qquad (9)$$

where $m(x,y)$ and $\delta(x,y)$ are the local mean and standard deviation of the pixels inside the local window and $k$ is a bias. The result is satisfactory at $k = -0.2$ and $w=15$. The local mean $m(x,y)$ and standard deviation $\delta(x,y)$ adapt the value of the threshold according to the contrast in the local neighborhood of the pixel. The bias $k$ controls the level of adaptation varying the threshold value.

#### 3.1.2 Sauvola's Technique

In Sauvola's binarization method, the threshold $T(x,y)$ is calculated using the mean $m(x,y)$ and standard deviation $\delta(x,y)$ of the pixels within a window of size $w \times w$ as:

$$T(x,y) = m(x,y)\left[1 + k\left(\dfrac{\delta(x,y)}{R} - 1\right)\right] \qquad (10)$$

where R is the maximum value of the standard deviation (R = 128 for a grayscale document), and $k$ is a bias, which takes positive values in the range [0.2, 0.5].

The local mean $m(x,y)$ and standard deviation $\delta(x,y)$ adapt the value of the threshold according to the contrast in the local neighborhood of the pixel. When there is high contrast in some region of the image, $s(x,y) \sim R$ which, results in $T(x,y) \sim m(x,y)$. This is the same result as in Niblack's method. However, the difference comes in when the contrast in the local neighbourhood is quite low. In that case the threshold $T(x,y)$ goes below the mean value thereby successfully removing the relatively dark regions of the background. The parameter $k$ controls the value of the threshold in the local window such that the higher the value of $k$, the lower the threshold from the local mean. A value of $k = 0.5$ was used by Sauvola1 and Sezgin. Badekas et al. Experiments with different values of k shows that $k = 0.34$ gives the best results. In general, the algorithm is not very sensitive to the value of $k$. The statistical constraint in Eq.(10) gives very good results even for severely degraded documents. However in order to compute the threshold $T(x,y)$, local mean and standard deviation have to be computed for each pixel. Computation of $m(x,y)$ and $\delta(x,y)$ in a naive way results in a computational complexity of $O(n^2w^2)$ for an image of size $n \times n$.

### 3.2 Local gray range method

J.Bernsen [14] uses the local gray range technique. In this technique the range between the maximum and minimum pixel gray range within the local window is used to determine the threshold value.

### 3.2.1 Bernsen's Technique:

In this method the local threshold value $T(x,y)$ at $(x,y)$ is calculated within a window of size $w \times w$ as:

$$T(x, y) = 0.5(I_{\max(i,j)} + I_{\min(i,j)}) \quad (11)$$

where $I_{max(i,j)}$ and $I_{min(i,j)}$ are maximum and minimum gray values within the local window, provided contrast

$$C(i, j) = I_{\max(i,j)} - I_{\min(i,j)} \geq 15 \quad (12)$$

In this method, the threshold is set at the midrange value, which is the mean of the maximum and minimum gray values in a local window of size $w \times w$. A value of $w = 31$ gives the satisfactory results. However, if the contrast $C(i,j)$ is below a certain threshold (15), then that neighborhood is said to consist of only one class, foreground or background, depending on the value of $T(x,y)$. There is no bias to control the threshold value.

## 4. Proposed Technique

To minimise the computational time of local thresholding calculation, we propose an efficient way of determining local threshold by using Eq.(7) and (13). Binarization can be speed up considerably as a result of using the integral sum image in Eq.(2-4) to determine the local sum in Eq.(6) where computational time does not depend on the window dimension. In other techniques like Sauvola's and Niblack's technique, local mean $m(x,y)$ and standard deviation $\delta(x,y)$ are required to determine the value of the threshold for each pixel. In this proposed technique, no local standard deviation is required and it requires to compute the local mean and mean deviation $\partial(x,y)$ to determine the local threshold as:

$$T(x, y) = m(x, y)\left[1 + k\left(\frac{\partial(x, y)}{1 - \partial(x, y)} - 1\right)\right] \quad (13)$$

where $\partial(x,y) = I(x,y) - m(x,y)$ is the local mean deviation and $k$ is a bias which can control the level of adaptation varying threshold value. Its range is [0,1] only.

Depending on the original scanned image type, the value of $k$ takes a major role in determining threshold value. The lower value of $k$ makes the threshold value higher and higher value of k lowers the threshold value. If the region within the window size is uniform, i.e. $I(x,y) = m(x,y)$ and then $\partial(x,y)$ becomes 0. In this case, the threshold value $T(x,y)$ be lower than mean $m(x,y)$ resulting the pixel $I(x,y)$ becomes background (white). If mean $m(x,y)$ is 0, then $T(x,y)$ also becomes 0 and the pixel $I(x,y)$ becomes background (white). If $k=0$ threshold $T(x,y)$ becomes the local mean $m(x,y)$. For different values of $k$, the adaptation level of threshold value can be adjusted.

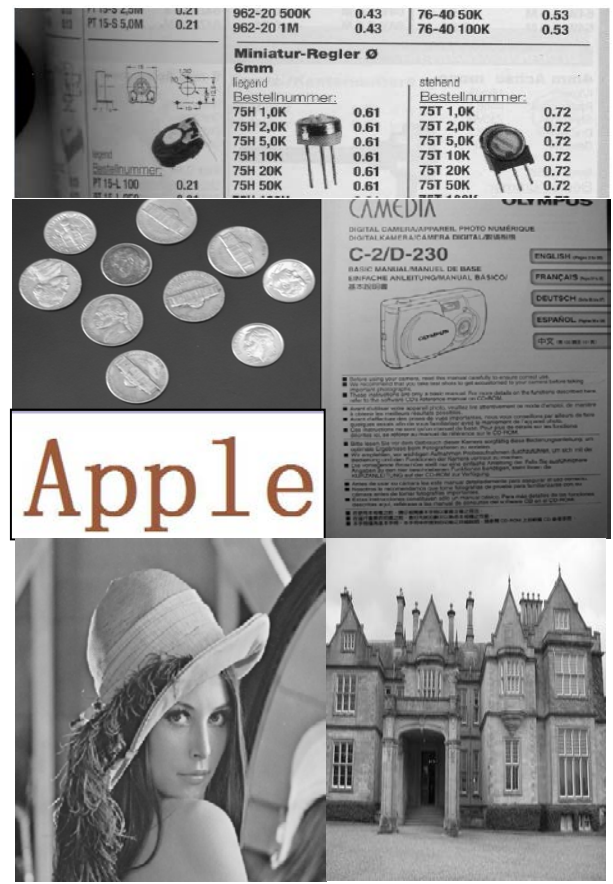

Fig. 2: Experiment Images.

Integral sum image is determined as a prior process for determining the local sum easily. If local sum is available, mean can be calculated with a simple arithmetic operation without depending on the window size as in Eq.(7). Calculation of local mean deviation is straightforward by just subtracting the mean from the concerned pixel. Thus this proposed technique can binaries faster than others.

Table1 : Computational Time Comparison on Lena image (in Sec)

| Window Size | Proposed | Bernsen | Niblack | Sauvola |
|---|---|---|---|---|
| 3 | 0.2496 | 0.8112 | 7.176 | 7.1448 |
| 7 | 0.234 | 1.3728 | 7.3944 | 7.3944 |
| 11 | 0.234 | 2.2152 | 7.9093 | 7.9561 |
| 15 | 0.234 | 3.4164 | 8.5177 | 8.5489 |
| 19 | 0.234 | 4.7892 | 9.2509 | 9.2821 |
| 23 | 0.234 | 6.3336 | 10.0777 | 10.0621 |
| 27 | 0.2184 | 8.1277 | 11.1073 | 11.0605 |
| 31 | 0.2028 | 9.9997 | 12.0745 | 12.1369 |
| 35 | 0.1872 | 12.0589 | 13.2913 | 13.3225 |

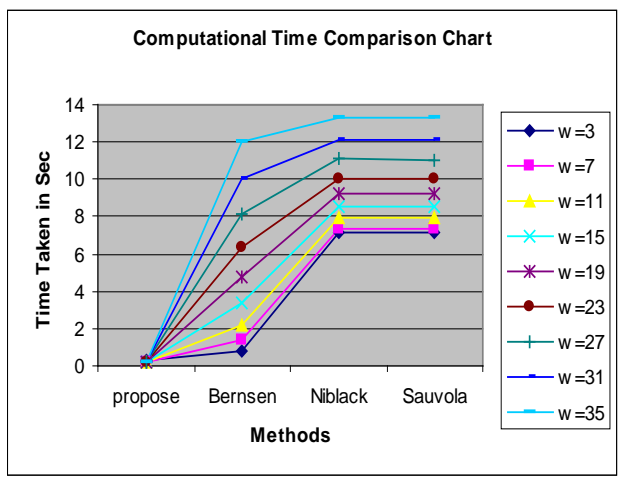

Fig. 3 Graphical Computational time Comparison for binarisation of different methods on image **lena**$_{512 \times 512}$.

## 5. Experimental Result

The experiment for the proposed system was carried out using MATLAB 7.3 (R2006b) on a PC with the following configuration: Intel® Core™2Duo CPU E6550,GHz 2.33 GHz, 2GB RAM, 32 bit OS(Windows Vista). Both qualitative and quantitative analysis are carried out in comparison of the proposed technique with other relevant techniques. Qualitative analysis provides a set of images for comparison to the reader for personal analysis. Proposed binarization technique was tested on different images and compared with several common binarization techniques such as Sauvola, Niblack and Bernsen. In figure 2 shows the set of images used in comparison. Comparison of computational times of the proposed technique with other local adaptive techniques is shown in figure 3 and Table 1.

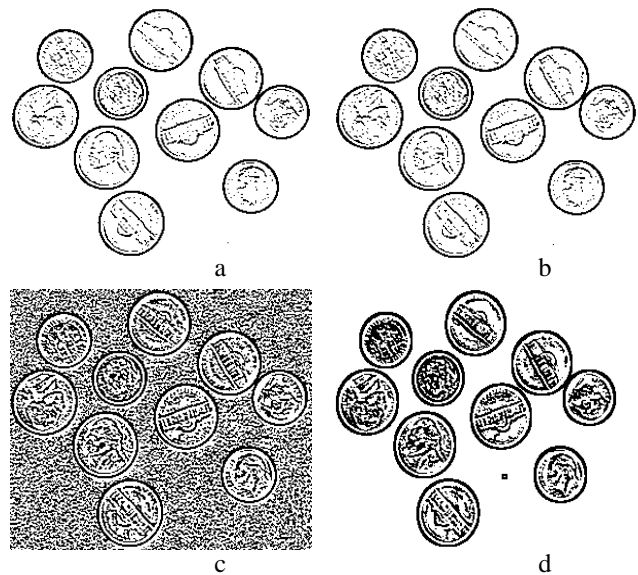

Fig.4 Comparison of various methods on coin image at 5×5 window size: (a) Proposed at $k=0.06$, (b) Sauvola's at $k=0.06$, (c) Niblack's at $k=-2$ and (d) Bernsen's.

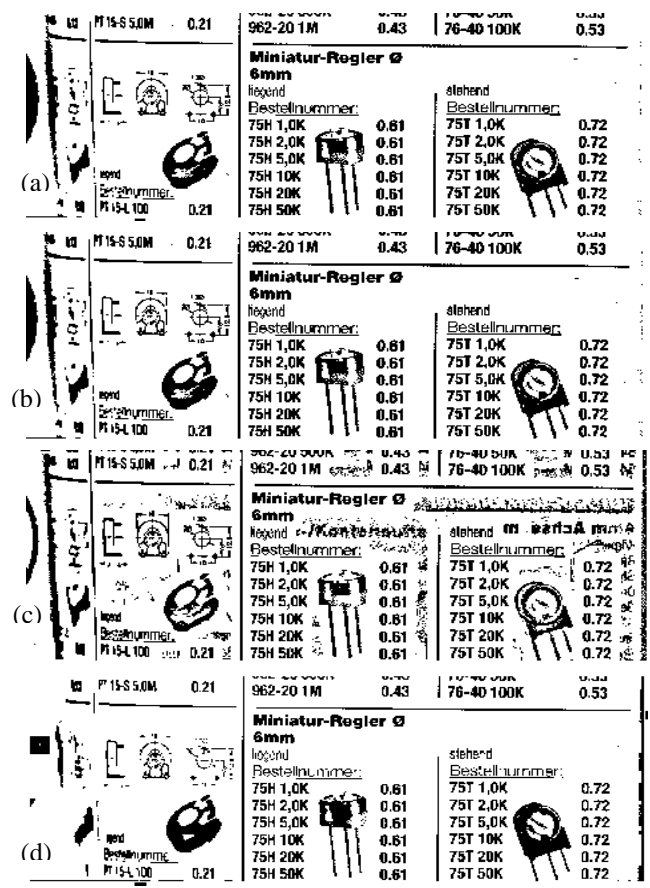

Fig. 5 Comparison on scanned document at 15×15 window size: (a) Proposed at $k=0.06$, (b). Sauvola's at $k=0.06$, (c). Niblack's at $k=-2$ (d). Bernsen's.

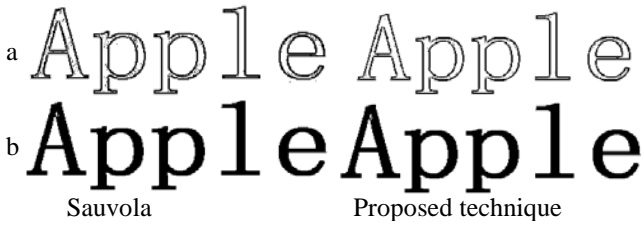

Fig.6: Experimental result on large text image at different window size: (a) Sauvola's at $k=5$ and (b) Proposed method at $k=15$.

In figures 4-8 show the test results of various methods for comparison. Performance varies at different window size. Niblack's and Bernsen's methods require large window size. For certain images, these two techniques are not suitable at smaller window size, for example at $5\times5$ while the proposed and Sauvola's methods give very good results, which are shown in figures 4 and 7. The proposed method gives good result even at very small window size. For document images, the size $w\times w$ is required to change depending on the character size. Otherwise only the character boundary is detected for large character size as shown in figure 6. But in this proposed method the computational time doesn't depend on window size while other methods depend on the window size. Its comparison chart and table are shown in figure 3 and Table1. Hence other methods has computational time complexity $O(w^2 \times n^2)$ for an $n\times n$ image while proposed technique has $O(n^2)$ only.

The motivation of this proposed system is to get a good binarised image without depending on the local window size as normally done in global techniques.

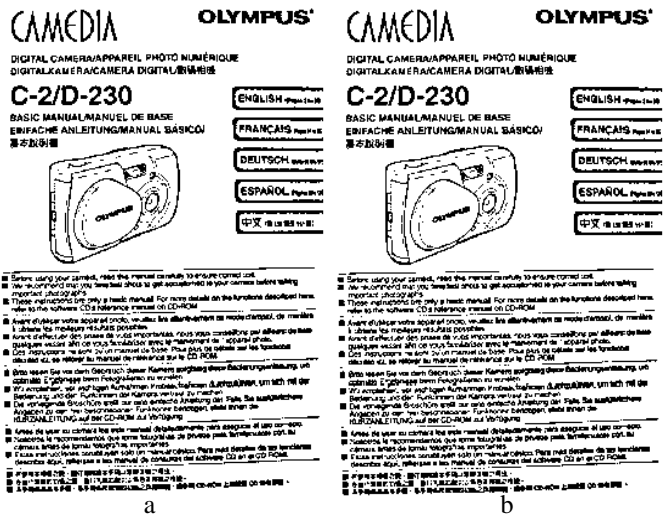

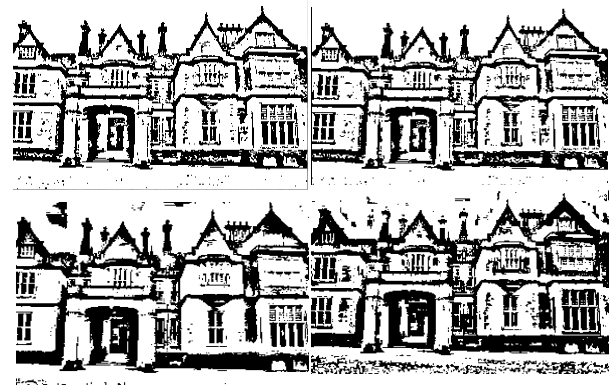

Fig. 8 Comparison on nondocument image at $15\times15$ window size: (a) Proposed at $k=0.06$, (b). Sauvola's at $k=0.06$, (c). Niblack's at $k=-2.5$ and window size $31\times31$ (d). Bernsen's.

Fig. 7 Comparison of various method on text document: (a) Proposed method at $k=0.06$ (b). *Sauvola's* at $k=0.06$ (c) Niblack's at $k=-2$ and (d) Bernsen's.

Comparison was carried out on both non document and document images. Mehmet Sezgin and Sankur [26] presented a comparative analysis of these methods and found that Sauvola's method is the best on non documental images, but somewhat poor in document images.

## 6. Conclusions

In this paper, we present a new way of computing thresholds for locally adaptive binarization scheme. We use integral sum image so that the running time does not depend on the local window size to compute mean in local windows. The proposed method is faster than Sauvola's method, and its running time approaches that of global binarization method. The proposed technique is compared with other relevant methods and found to be better than other contemporary methods, both in terms of quality and speed.

**REFERENCES**

[1] M. Kamel and A. Zhao, ''Extraction of binary character/graphics images from grayscale document images,'' Graph. Models Image Process.55(3), 203–217(1993).

[2] T. Abak, U. Baris, and B. Sankur, ''The performance of thresholding algorithms for optical character recognition,'' Intl. Conf. Document Anal. Recog. ICDAR'97, pp. 697–700 (1997).

[3] O.D. Trier and A. K. Jain, ''Goal-directed evaluation of binarization methods,'' IEEE Trans. Pattern Anal. Mach. Intell. PAMI-17, 1191– 1201(1995).

[4] B. Bhanu, ''Automatic target recognition: state of the art survey,'' IEEE Trans. Aerosp. Electron. Syst. AES-22, 364–379 (1986).

[5] M. Sezgin and R. Tasaltin, ''A new dichotomization technique to multilevel thresholding devoted to inspection applications,'' Pattern Recogn. Lett. 21, 151–161 (2000).

[6] M. Sezgin and B. Sankur, ''Comparison of thresholding methods for non-destructive testing applications,'' IEEE ICIP'2001, Intl. Conf. Image Process., pp. 764–767 (2001).

[7] J. Sauvola and M. Pietikainen, "Adaptive document image binarization," Pattern Recognition 33(2), pp. 225–236, 2000.

[8] N. Otsu, "A threshold selection method from gray-level histograms," IEEE Trans. Systems, Man, and Cybernetics 9(1), pp. 62–66, 1979.

[9] P. Viola and M. J. Jones, "Robust real-time face detection," Int. Journal of Computer Vision 57(2), pp. 137– 154, 2004.

[10] R. Cattoni, T. Coianiz, S. Messelodi, and C. M.Modena, "Geometric layout analysis techniques for document image understanding: a review," tech. rep., IRST, Trento, Italy, 1998.

[11] F. Shafait, D. Keysers, and T. M. Breuel, "Performance comparison of six algorithms for page segmentation," in 7th IAPR Workshop on Document Analysis Systems, pp. 368–379, (Nelson, New Zealand), Feb. 2006.

[12] J. M. White and G. D. Rohrer, "Image thresholding for optical character recognition and other applications requiring character image extraction," IBM Journal of Research and Development 27, pp. 400–411, July 1983.

[13] Chi, Z., Yan, H., and Pham, T.: 'Fuzzy algorithms: with applications to image processing and pattern recognition' (World Scientific Publishing, 1996)

[14] Bernsen, J.: 'Dynamic thresholding of gray-level images'. Proc. 8th Int. Conf. on Pattern Recognition, Paris, 1986, pp. 1251–1255

[15] Chow, C.K., and Kaneko, T.: 'Automatic detection of the left ventricle from cineangiograms', Comput. Biomed. Res., 1972, 5, pp. 388–410

[16] Eikvil, L., Taxt, T., and Moen, K.: 'A fast adaptive method for binarization of document images'. Proc. ICDAR, France, 1991, pp. 435–443

[17] Mardia, K.V., and Hainsworth, T.J.: 'A spatial thresholding method for image segmentation', IEEE Trans. Pattern Anal. Mach. Intell., 1988, 10, (8), pp. 919–927

[18] Niblack, W.: 'An introduction to digital image processing' (Prentice- Hall, Englewood Cliffs, NJ, 1986), pp. 115–116

[19] Taxt, T., Flynn, P.J., and Jain, A.K.: 'Segmentation of document images', IEEE Trans. Pattern Anal. Mach. Intell., 1989, 11, (12), pp. 1322–1329

[20] Yanowitz, S.D., and Bruckstein, A.M.: 'A new method for image segmentation', Comput. Vis. Graph. Image Process., 1989, 46, (1), pp. 82–95

[21] Sauvola, J., Seppanen, T., Haapakoski, S., and Pietikainen, M.: 'Adaptive document binarization'. Proc. 4th Int. Conf. on Document Analysis and Recognition, Ulm Germany, 1997, pp. 147–152

[22] Gorman, L.O.: 'Binarization and multithresholding of document images using connectivity', CVGIP, Graph. Models Image Process., 1994, 56, (6), pp. 494–506

[23] Liu, Y., and Srihari, S.N.: 'Document image binarization based on texture features', IEEE Pattern Anal. Mach. Intell., 1997, 19, (5), pp. 540–544

[24] W. Niblack, An Introduction to Image Processing, Prentice-Hall, Englewood Cliffs, NJ, 1986.

[25] L. O'Gorman, "Binarization and multithresholding of document images using connectivity," Graphical Model and Image Processing 56, pp. 494–506, Nov. 1994.

[26] Mehmet Sezgin and Bülent Sankur "Survey over image thresholding techniques and quantitative performance evaluation", Journal of Electronic Imaging 13(1), 146–165 (January 2004).

[27] F. C. Crow, "Summed-area tables for texture mapping," Computer Graphics - Proceedings of SIGGRAPH'84 18(3), pp. 207–212, 1984.

[28] B. Gatos, I. Pratikakis and S.J. Perantonis,' Improved Document Image Binarization by Using a Combination of Multiple Binarization Techniques and Adapted Edge Information', 978-1-4244-2175-6/08/$25.00 ©2008 IEEE

[29] B. Su, S. Lu, and C. L. Tan, "Document Image Binarization Using Background Estimation and Stroke Edges," *Proc. Intl. Journal on Document Analysis &Recognition*, Vol. 13, No. 4, pp. 303-314, 2010.

[30] O. H. Hwa, L. T. Kil, and C. I. Sung, "An Improved Binarization Algorithm Based on a Water Flow Model for Document Image with Inhomogeneous Backgrounds," *Pattern Recognition*, Vol. 38, pp. 2612 –2625, 2005.

[31] B. Su, S. Lu, and C. L. Tan, "Binarization of Historical Document Images Using The Local Maximum and Minimum," Proc. Intl. Workshop on Document Analysis Systems, pp. 159–165, June 2010.

[32] B. Gatos, I. Pratikakis, and S. J. Perantonis, "Adaptive Degraded Document Image Binarization," Pattern Recognition, Vol. 39, No. 3, pp. 317–327, March 2006.

[33] T.Hoang Ngan Le, Tien D. Bui, and Ching Y. suen, "Ternary Entropy – based Binarisation of Degraded Document Images Using Morphological Operators" 1520-5363/11$26.00 ©2011 IEEE.